\documentclass{article}
\usepackage{spconf,amsmath,graphicx,hyperref}

\usepackage{graphicx}
\usepackage{amsmath}
\usepackage{amssymb}
\usepackage{booktabs}

\usepackage{xcolor} 
\usepackage{times}
\usepackage{epsfig}
\usepackage{graphicx}
\usepackage{amsmath}
\usepackage{amssymb}
\usepackage[bb=dsserif]{mathalpha}
\usepackage{bm}

\usepackage{booktabs,colortbl,tabularx}
\usepackage{pifont}%

\usepackage{mathtools}
\usepackage{commath}
\usepackage{algorithm}
\usepackage{algpseudocode}

\usepackage{multirow}
\usepackage{comment} 
\usepackage{enumitem}
\usepackage{minted}

\definecolor{Gray}{gray}{0.9}

\usepackage{xspace}
\newcommand{\method}{{\selectfont\emph{\textsc{GMS-CAVP}}}\xspace}

\newcommand{\cmark}{\ding{51}}%
\newcommand{\xmark}{\ding{55}}%

\definecolor{battleshipgrey}{rgb}{0.52, 0.52, 0.51}

\title{\method: Improving Audio-Video Correspondence with Multi-Scale Contrastive and Generative Pretraining}
\name{Shentong Mo$^{1,2,3,\dagger}$, Zehua Chen$^{3,\dagger}$, Jun Zhu$^{3}$\sthanks{Corresponding author, $^\dagger$ Equal contribution.}}
\address{$^1$Carnegie Mellon University, $^2$MBZUAI, $^3$Tsinghua University}
\begin{document}
%
\maketitle
\begin{abstract}

Recent advances in video-audio (V-A) understanding and generation have increasingly relied on joint V-A embeddings, which serve as the foundation for tasks such as cross-modal retrieval and generation. 
While prior methods like CAVP effectively model semantic and temporal correspondences between modalities using contrastive objectives, their performance remains suboptimal. 
A key limitation is the insufficient modeling of the dense, multi-scale nature of both video and audio signals, correspondences often span fine- to coarse-grained spatial-temporal structures, which are underutilized in existing frameworks.
To this end, we propose \method, a novel framework that combines Multi-Scale Video-Audio Alignment and Multi-Scale Spatial-Temporal Diffusion-based pretraining objectives to enhance V-A correspondence modeling. 
First, \method\ introduces a multi-scale contrastive learning strategy that captures semantic and temporal relations across varying granularities. 
Second, we go beyond traditional contrastive learning by incorporating a diffusion-based generative objective, enabling modality translation and synthesis between video and audio. 
This unified discriminative-generative formulation facilitates deeper cross-modal understanding and paves the way for high-fidelity generation.
Extensive experiments on VGGSound, AudioSet, and Panda70M demonstrate that \method\ outperforms previous methods in generation and retrieval.

\end{abstract}


\section{Introduction}

Learning robust video-audio (V-A) correspondence lies at the heart of many cross-modal tasks~\cite{Iashin2021SpecVQGAN,sheffer2023im2wav,yang2022diffsound,kreuk2023audiogen,liu2023audioldm}, including video-to-audio generation, audio-driven video synthesis, and audio-visual retrieval. In particular, recent efforts have focused on contrastive audio-visual pretraining (CAVP)~\cite{luo2023difffoley,liu2023audioldm} to align modalities via discriminative objectives. These approaches have demonstrated impressive progress by embedding video and audio into a shared space, enabling downstream retrieval tasks.

Recent approaches~\cite{Iashin2021SpecVQGAN, sheffer2023im2wav, kreuk2023audiogen, luo2023difffoley} to video-to-audio generation primarily rely on contrastive pretraining to improve modality alignment and diffusion-based generative models to synthesize high-quality audio. Contrastive learning techniques project video and audio data into a shared latent space, enhancing the relevance of generated audio. However, these methods often struggle to capture the fine-grained, multi-scale spatial-temporal dependencies necessary for effective cross-modal understanding and high-fidelity audio generation. The lack of hierarchical feature modeling limits their capacity to generalize across diverse video scenes, particularly those involving complex motion patterns or intricate audio-visual relationships.

However, two core limitations remain underexplored. First, although CAVP serves as a backbone for both retrieval and generation tasks, current pretraining objectives are exclusively contrastive in nature. This neglects the inherent modality translation capabilities required in generative tasks such as video-to-audio synthesis. Consequently, models pretrained solely with discriminative losses are suboptimal when deployed in generative settings, where learning cross-modal mappings is crucial. Second, both video and audio are information-dense modalities that span multiple spatial and temporal scales. Yet existing CAVP methods apply a single-scale global alignment strategy, which overlooks fine-grained and hierarchical cross-modal correspondences, essential for capturing rich V-A dynamics.

To address these challenges, we propose \method, a novel multi-scale contrastive and generative framework for cross-modal pretraining. \method\ introduces a Multi-scale Video-Audio Alignment mechanism to enforce hierarchical video-audio correspondence and a Generative Multi-scale Video-Audio Alignment to bridge the generative gap between video and audio modalities. By integrating contrastive learning with diffusion-based generative modeling, our approach ensures robust cross-modal alignment while significantly improving the fidelity and coherence of generated audio. Unlike prior methods that operate on single-scale representations, \method\ leverages hierarchical spatial-temporal structures to enhance synchronization and realism.

We conduct extensive experiments on VGGSound, AudioSet, and Panda70M to validate the effectiveness of \method. Our results demonstrate that \method\ achieves state-of-the-art performance in both video-to-audio generation and retrieval tasks. \method\ consistently outperforms existing approaches in key metrics such as KLD, FAD, and Align Acc, confirming its ability to generate high-quality, temporally coherent audio while maintaining precise synchronization with visual inputs.
Ablation studies further show the importance of our proposed components, and explore the effect of diffusion sampling, bidirectional training, spatial multi-scale, and data scaling.

We summarize our contributions below:
\begin{itemize}
    \item We propose \method, a unified multi-scale contrastive and generative approach to learn cross-modal correspondence
    for generation and retrieval.  
    \item We introduce Multi-scale Spatial-temporal Alignment and Multi-scale Spatial-temporal Diffusion for video-audio pretraining. 
    \item We conduct extensive experiments and ablation studies demonstrating that \method\ outperforms existing methods in generation and retrieval.  
\end{itemize}

\section{Related Work}

\noindent{\textbf{Video-Audio Correspondence.}}  
Video-to-audio generation~\cite{Iashin2021SpecVQGAN,sheffer2023im2wav,luo2023difffoley,mei2023foleygen}, the task of translating visual information into corresponding audio outputs, has seen rapid advancements driven by model architectures.  
Recent works have explored multimodal pre-training and fine-tuning strategies. Seeing \& Hearing~\cite{xing2024seeing} utilizes the pre-trained ImageBind model~\cite{girdhar2023imagebind} as a latent aligner for cross-modal diffusion-based generation. VAB~\cite{su2024vision} introduces a visual-conditioned masked audio token prediction task for pre-training without diffusion models. MaskVAT~\cite{pascual2024masked} combines a sequence-to-sequence masked generative model with a high-quality audio codec to achieve temporal synchronicity. VATT~\cite{liu2024tell} integrates large language models like Gemma-2B~\cite{gemmateam2024gemma} and LLaMA-2-7B~\cite{touvron2023llama} with projection layers to map video features into audio tokens.  
In contrast, \method\ introduces a unified multi-scale contrastive and generative framework, leveraging multi-scale spatial-temporal alignment to improve video-audio correspondence across different temporal and spatial resolutions. Unlike prior works that focus solely on global representations, \method\ ensures robust alignment through multi-scale contrastive learning and further enhances generation quality via multi-scale spatial-temporal diffusion.

\noindent{\textbf{Video-Audio Generative Modeling.}}  
Recent advancements~\cite{kreuk2023audiogen,yang2022diffsound} have explored diffusion-based generative models for cross-modal learning. 
In video-audio generation, Diff-Foley~\cite{luo2023difffoley} integrates contrastive pre-training with diffusion-based decoding to synthesize synchronized audio from visual inputs.  
Our work builds upon these developments by introducing multi-scale spatial-temporal diffusion, which enhances video-conditioned audio generation using a hierarchical, multi-scale conditioning mechanism. Unlike previous methods that operate on a single-level feature representation, \method\ progressively refines multi-scale video features, improving both fidelity and synchronization in generated audio.

\section{Method}

In this section, we present the proposed multi-scale discriminative and generative framework, namely \method, which enables contrastive audio-video pretraining for learning both discriminative and generative representations. We first provide preliminaries by defining the problem setup. Then, we introduce the \textit{Multi-scale Spatial-temporal Alignment} mechanism, which captures fine-grained hierarchical dependencies for enhanced video-audio correspondence. Finally, we propose \textit{Multi-scale Spatial-temporal Diffusion} to bridge the generative gap between video and audio representations for improved video-to-audio synthesis.

\begin{figure*}[t]
\centering
\includegraphics[width=0.7\linewidth]{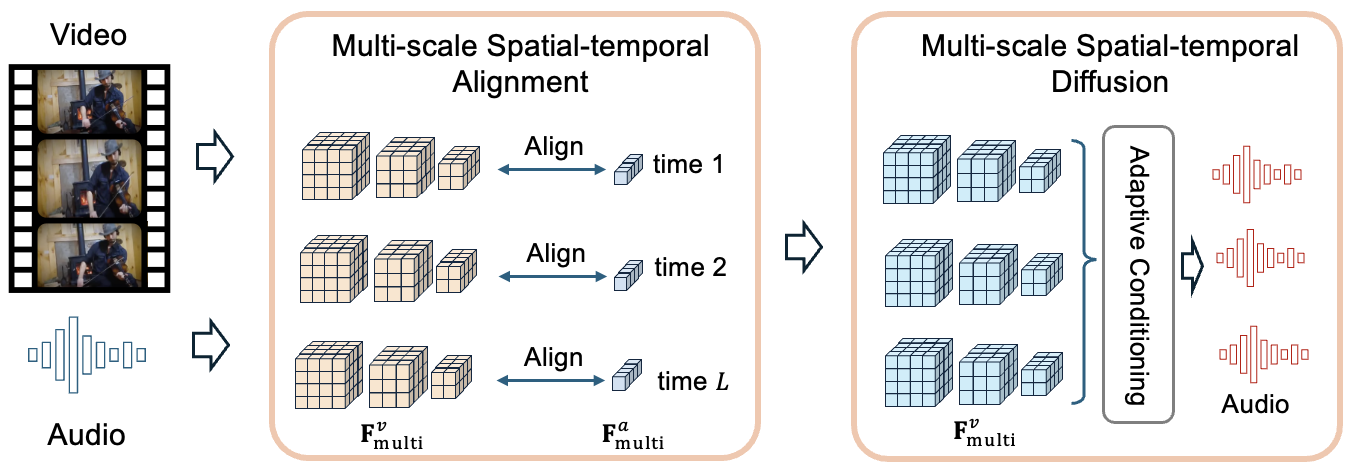}
\vspace{-1.0em}
\caption{Illustration of the proposed multi-scale discriminative and generative architecture (\method) for learning audio-video correspondence. 
We introduce the \textit{Multi-scale Video-Audio Alignment} mechanism, which captures fine-grained hierarchical dependencies for enhanced video-audio correspondence. 
Then, \textit{Generative Multi-scale Video-Audio Alignment} is proposed to bridge the generative gap between video and audio representations for improved video-to-audio synthesis.
}
\label{fig: main_img}
\vspace{-1.0em}
\end{figure*}

\subsection{Preliminaries}\label{sec: pre}

In this section, we describe the problem setup for video-audio pretraining.

\noindent\textbf{Problem Setup and Notations.}
Given audio $a$ and visual frames $v$ from a video, the goal is to bridge the cross-modal gap between audio and video/text learned in a contrastive space. For a given video, the corresponding mel-spectrogram of the audio is represented as $\mathbf{A} \in \mathbb{R}^{T\times F}$, where $T$ and $F$ denote the time and frequency dimensions, respectively. The visual frames are denoted as $\mathbf{V} \in \mathbb{R}^{T\times H \times W \times 3}$, where $H$ and $W$ represent the spatial resolution.
Each modality is processed using a pre-trained encoder: $f_a(\cdot)$ for audio, $f_v(\cdot)$ for video, yielding feature representations:
\begin{equation}
    \mathbf{F}^a = f_a(\mathbf{A}), \quad
    \mathbf{F}^v = f_v(\mathbf{V}).
\end{equation}
These embeddings are projected into a shared feature space to facilitate contrastive and generative learning.

\subsection{Multi-scale Video-Audio Alignment}\label{sec: msa}

One of the key challenges in video-to-audio generation is ensuring that the generated audio remains temporally synchronized and semantically aligned with the corresponding visual content. Traditional contrastive learning approaches primarily rely on global feature representations, which often fail to capture fine-grained spatial and temporal dependencies. This leads to suboptimal cross-modal alignment, particularly in scenarios involving complex motions, rapid scene transitions, or intricate audio variations.

To address this, we propose a Multi-scale Spatial-temporal Alignment (MSA) mechanism that captures hierarchical dependencies across multiple resolutions, as illustrated in Figure~\ref{fig: main_img}. Unlike existing methods that enforce alignment only at a single temporal or spatial scale, our approach learns structured representations at varying granularities, ensuring both local and global coherence. 

\noindent\textbf{Hierarchical Representation Learning.}  
Instead of treating the video and audio as monolithic signals, we construct multi-scale embeddings to capture relationships at different levels of abstraction. Given the extracted video feature sequence $\mathbf{F}^{v}$ and its corresponding audio feature sequence $\mathbf{F}^{a}$, we decompose them into multiple resolutions:
\begin{equation}
    \mathbf{F}^{v}_{\text{multi}} = \{\mathbf{F}^{v}_{1}, \mathbf{F}^{v}_{2}, \dots, \mathbf{F}^{v}_{L} \}, \quad
    \mathbf{F}^{a}_{\text{multi}} = \{\mathbf{F}^{a}_{1}, \mathbf{F}^{a}_{2}, \dots, \mathbf{F}^{a}_{L} \},
\end{equation}
where $\mathbf{F}^{v}_{l}$ and $\mathbf{F}^{a}_{l}$ represent features extracted at different spatial-temporal scales $l=1, \dots, L$. These hierarchical features are obtained via a combination of temporal pyramidal pooling and multi-resolution convolutions, ensuring that both short-term and long-term dependencies are captured effectively.

\noindent\textbf{Multi-scale Contrastive Learning.}  
To enhance alignment at different scales, we extend the contrastive objective to operate at multiple resolutions. For each scale $l$, we compute the similarity between video and audio features using cosine similarity:
\begin{equation}
    \mathtt{sim}(\mathbf{F}^{a}_{l}, \mathbf{F}^{v}_{l}) = \frac{\mathbf{F}^{a}_{l} \cdot \mathbf{F}^{v}_{l}}{\|\mathbf{F}^{a}_{l}\| \|\mathbf{F}^{v}_{l}\|}.
\end{equation}
The contrastive loss is then formulated as:
\begin{equation}
    \mathcal{L}_{\text{MSA}} = \sum_{l=1}^{L} \mathcal{L}_{\text{CL}} (\mathbf{F}^{a}_{l}, \mathbf{F}^{v}_{l}),
\end{equation}
where $\mathcal{L}_{\text{CL}}$ is the standard InfoNCE contrastive loss.
This encourages cross-modal features to be maximally aligned at each scale, improving fine-grained synchronization between audio and video.

\noindent\textbf{Adaptive Temporal Alignment.}  
To further refine alignment, we introduce an adaptive temporal alignment mechanism that applies an attention-based weighting scheme to emphasize salient temporal regions:
\begin{equation}
    \mathbf{w}_t = \mathtt{softmax}(\mathbf{F}^{v}_{t} \cdot \mathbf{F}^{a}_{t}),
\end{equation}
where $\mathbf{w}_t$ represents an importance weight vector that modulates the contribution of each temporal segment in the final contrastive objective. This ensures that key moments in the video (e.g., action frames, object interactions) receive higher alignment emphasis, mitigating the impact of irrelevant or noisy segments.

\vspace{-0.5em}

\subsection{Generative Multi-scale Video-Audio Alignment}\label{sec: msdiff}

While contrastive learning ensures effective video-audio alignment, it does not directly model the generative relationship between the two modalities. Video and audio have inherently different structures where video features tend to be high-dimensional and spatially structured, whereas audio is sequential and exhibits strong frequency dependencies. This discrepancy poses a fundamental challenge for generating realistic and synchronized audio from video inputs.

To address this, we introduce a Multi-scale Spatial-temporal Diffusion model that progressively refines audio representations conditioned on hierarchical video features. Our approach leverages a diffusion-based decoder trained to bridge the modality gap by denoising latent representations in a structured manner.

\noindent\textbf{Diffusion-based Audio Generation.}  
Given a video feature sequence $\mathbf{F}^v_{\text{multi}}$, we define the generative process as:
\begin{equation}
    p_{\boldsymbol{\theta}}(\mathbf{A}_0 | \mathbf{V}) = \prod_{t=1}^{T} p_{\boldsymbol{\theta}}(\mathbf{A}_t | \mathbf{A}_{t+1}, \mathbf{F}^v_{\text{multi}}),
\end{equation}
where each step conditions on the video features across multiple scales. This formulation allows the model to generate temporally coherent audio by leveraging video dynamics at different resolutions.

To achieve stable and high-quality generation, we employ a hierarchical diffusion model where noise is gradually removed from the audio representation at multiple resolutions:

\begin{equation}
\begin{aligned}
    \mathcal{L}_{\text{MSD}} & = \mathbb{E}_{t, \mathbf{A}_0, \boldsymbol{\epsilon}} \left[ \|\boldsymbol{\epsilon} - \boldsymbol{\epsilon}_{\boldsymbol{\theta}}(\mathbf{A}_{t}, t, \mathbf{F}^v_{\text{multi}})\|^2 \right] \\
    & + 
    \mathbb{E}_{t, \mathbf{A}_T, \boldsymbol{\epsilon}} \left[ \|\boldsymbol{\epsilon} - \boldsymbol{\epsilon}_{\boldsymbol{\theta}}(\mathbf{A}_{T}, T, \mathbf{F}^v_{\text{multi}})\|^2 \right].
\end{aligned}
\end{equation}

\section{Experiments}

\subsection{Experimental Setup}

\noindent \textbf{Datasets.}
We conduct experiments on three large-scale datasets: 
VGGSound~\cite{chen2020vggsound} consists of 200k YouTube video clips, each lasting 10 seconds, covering 309 diverse sound categories such as animal sounds, vehicles, human speech, musical instruments, and environmental noises.  
AudioSet~\cite{Gemmeke2017audioset} is a large-scale audio-visual dataset with approximately 2M YouTube videos spanning a wide range of real-world sounds.  
Panda70M~\cite{chen2024panda70m} is a newly introduced large-scale audio-visual corpus containing 70M video-audio pairs.

\noindent \textbf{Evaluation Metrics.}
To assess the quality of video-to-audio generation, we employ the following metrics:
Kullback-Leibler Divergence (KLD) measures the distributional similarity between generated and ground-truth (GT) audio features based on PaSST representations~\cite{koutini2022efficient}, evaluating semantic coherence.
Fréchet Audio Distance (FAD)~\cite{kilgour2018frechet} quantifies the quality of generated audio by comparing feature distributions of real and generated samples.
Alignment Accuracy (Align Acc)~\cite{luo2023difffoley} evaluates the temporal consistency between generated audio and video frames, measuring synchronization performance.
For retrieval tasks, recall at rank $k$ (R@$k$, $k$ = 1, 5, 10) measures the percentage of labels retrieved within the top $k$ ranked predictions, and the higher value is better.

\noindent \textbf{Implementation.}
The input video frames are resized to $224 \times 224$ pixels. Audio inputs are log spectrograms extracted from 10-second clips sampled at 8000Hz. 
Audio spectrograms are computed using Short-Time Fourier Transform (STFT) with a 50ms window and a 25ms hop size, resulting in tensors of size $128 \times 128$ (128 frequency bins over 128 time steps).    
Training is performed for 200 epochs using the Adam optimizer~\cite{kingma2014adam} with a learning rate of $1e-4$ and a batch size of 64. 
We use the setting for diffusion generator in~\cite{luo2023difffoley}.

\begin{table}[t]
\renewcommand\tabcolsep{2.0pt}
\renewcommand{\arraystretch}{1.1}
\centering
\caption{Comparison results on video-to-audio generation. 
}
\label{tab: exp_gen}
\scalebox{0.78}{
\begin{tabular}{lccc}
\toprule
\bf Method            & \bf KLD $\downarrow$ & \bf FAD $\downarrow$ & \bf Align Acc $\uparrow$ \\ \midrule
SpecVQGAN~\cite{Iashin2021SpecVQGAN}         & 3.78 & 6.63 & 48.79 \\
Im2Wav~\cite{sheffer2023im2wav}            & 2.54 & 6.32 & 74.31 \\
Diff-Foley~\cite{luo2023difffoley}        & 3.15 & 6.40 & 82.47 \\
FoleyGen~\cite{mei2023foleygen}          & 2.89 & 2.59 & 73.83 \\
V2A-Mapper~\cite{wang2024v2amapper}        & 2.78 & 0.99 & 74.37 \\
Seeing \& Hearing~\cite{xing2024seeing} & 2.62 & 2.63 & 78.95 \\
MaskVAT~\cite{pascual2024masked}           & 2.65 & 1.51 & 63.87 \\
VAB~\cite{su2024vision}               & 2.58 & 2.69 & 76.83 \\
VATT~\cite{liu2024tell}              & 2.25 & 2.35 & 82.81 \\
\method\ (ours) & \bf 1.63 & \bf 0.75 & \bf 95.87 \\
\bottomrule
\end{tabular}}
\vspace{-0.5em}
\end{table}

\vspace{-0.5em}

\subsection{Comparison to Prior Work}\label{sec:exp}

In order to demonstrate the effectiveness of the proposed \method, we compare with a range of strong video-to-audio generation models~\cite{Iashin2021SpecVQGAN,sheffer2023im2wav,luo2023difffoley,mei2023foleygen,xing2024seeing,su2024vision,pascual2024masked,liu2024tell}.

Table~\ref{tab: exp_gen} reports a comprehensive comparison of \method\ against prior video-to-audio generation models on the VGGSound test set. Our model consistently outperforms previous approaches across all evaluation metrics, demonstrating superior cross-modal alignment and audio synthesis quality.

\begin{table}[!t]
    \renewcommand\tabcolsep{2.0pt}
    \renewcommand{\arraystretch}{1.1}
	\centering
        \caption{Comparison results of video-to-audio retrieval and audio-to-video retrieval.}
	\label{tab: exp_retrieval}
	\scalebox{0.7}{
	    \begin{tabular}{lccccccc}
			\toprule
			\multirow{2}{*}{\bf Method} & \multicolumn{3}{c}{\bf Video-to-Audio} & \multicolumn{3}{c}{\bf Audio-to-Video} \\
			& \bf R@1 & \bf R@5 & \bf R@10 & \bf R@1 & \bf R@5 & \bf R@10 \\
			\midrule
                CAVP~\cite{luo2023difffoley} &  9.50  & 25.40  & 35.10  & 11.10  & 27.80  & 36.40  \\
                \method\ (ours) & \bf 28.90 & \bf 43.70 & \bf 57.90 & \bf 30.50 & \bf 45.30 & \bf 58.20 \\
			\bottomrule
	    \end{tabular}}
     \vspace{-1.5em}
 \end{table}

\begin{table}[t]
\centering
\caption{Ablation studies on Multi-scale Spatial-temporal Alignment (MSA) and Multi-scale Spatial-temporal Diffusion (MSD) for video-to-audio generation and retrieval.}
\label{tab: ab_component}
\scalebox{0.7}{
\begin{tabular}{cccccccc}
\toprule
\bf MSA & \bf MSD  & \bf KLD $\downarrow$ & \bf FAD $\downarrow$ & \bf Align Acc $\uparrow$ & \bf R@1 & \bf R@5 & \bf R@10 \\ \midrule
\xmark & \xmark & 3.15 & 6.40 & 82.47 & 9.50  & 25.40 & 35.10 \\
\cmark & \xmark & 2.06 & 1.37 & 90.76 & 22.30 & 37.80 & 50.60 \\
\xmark & \cmark & 2.17 & 1.58 & 89.85 & 20.80 & 36.30 & 46.70 \\
\cmark & \cmark & \bf 1.63 & \bf 0.75 & \bf 95.87 & \bf 28.90 & \bf 43.70 & \bf 57.90 \\
\bottomrule
\end{tabular}}
\vspace{-1.5em}
\end{table}

To further validate \method\ in retrieval, we compare its performance against CAVP~\cite{luo2023difffoley} using R@$1$, R@$5$, and R@$10$ metrics. 
As shown in Table~\ref{tab: exp_retrieval}, \method\ significantly outperforms CAVP across all recall metrics, highlighting its improved video-audio representation learning.

\subsection{Experimental Analysis}

In this section, we performed ablation studies to demonstrate the benefits of Multi-scale Spatial-temporal Alignment (MSA) and Multi-scale Spatial-temporal Diffusion (MSD) pretraining objectives, as shown in Table~\ref{tab: ab_component}.
Furthermore, we conducted extensive experiments to explore the number of sampling steps, bidirectional diffusion training interval, the effect of spatial multi-scale, and data scaling of Multi-scale Spatial-temporal Alignment.
When training with VGGSound + AudioSet + Panda70M, the model achieves the best overall results, with the lowest KLD (1.35) and FAD (0.58), indicating the highest fidelity in audio generation.

\section{Conclusion}

In this work, we present \method, a novel framework for contrastive audio-video pretraining that unifies multi-scale discriminative and generative learning. 
By integrating Multi-scale Spatial-temporal Alignment and Multi-scale Spatial-temporal Diffusion pretraining objectives, our approach captures fine-grained video-audio correspondences and generative alignment. 
Extensive experiments on VGGSound, AudioSet, and Panda70M validate the effectiveness of \method, achieving state-of-the-art performance in both video-to-audio generation and retrieval. 
Ablation studies further confirm the importance of our proposed components.

\noindent\textbf{Acknowledgements.}
This work is supported by the National Natural Science Foundation of China (62550004, U24A20342, U25B6003, 92570001).

\vfill\pagebreak

\bibliographystyle{IEEEbib}
\bibliography{reference}

\end{document}